\documentclass{article}
\usepackage{spconf,amsmath,graphicx,bbm,mathtools}
\usepackage{multirow, hhline, stmaryrd, url, cases}

\DeclareMathOperator*{\argmin}{arg\,min}

\usepackage{hyperref}

\title{Adversarial Speaker Adaptation}
\name{Zhong Meng, Jinyu Li, Yifan Gong}
 \address{Microsoft Corporation, Redmond, WA, USA \\ \normalsize\{zhme, jinyl, ygong\}@microsoft.com}

\begin{document}
\ninept
\maketitle
\begin{abstract}


We propose a novel adversarial speaker adaptation (ASA) scheme, in which
adversarial learning is applied to regularize the distribution of
deep hidden features in a speaker-dependent (SD) deep neural network
(DNN) acoustic model to be close to that of a \emph{fixed}
speaker-independent (SI) DNN acoustic model during adaptation. 
An additional discriminator network is introduced
to distinguish the deep features generated by the SD model from those
produced by the SI model.  In ASA, with a fixed SI model as the
reference, an SD model is jointly optimized with the discriminator network
to minimize the senone classification loss, and simultaneously to 
mini-maximize the SI/SD discrimination loss on the adaptation data.
With ASA, a senone-discriminative deep feature is learned in the SD model with a
similar distribution to that of the SI model. With such a
regularized and adapted deep feature, the SD model can perform improved 
automatic speech recognition on the target speaker's speech. 
Evaluated on the Microsoft short message dictation dataset, ASA achieves 14.4\% and 7.9\%
relative word error rate improvements for supervised and unsupervised adaptation,
respectively, over an SI model trained from 2600 hours data, with 200 adaptation utterances per speaker. 

\end{abstract}

\section{Introduction}
\label{sec:intro}

With the advent of deep learning, the performance of automatic speech
recognition (ASR) has greatly improved \cite{DNN4ASR-hinton2012,
yu2017recent}. However, the ASR performance is not optimal when acoustic
mismatch exists between training and testing \cite{Li14overview}.
Acoustic model adaptation is a natural solution to compensate for this
mismatch.  For speaker adaptation, we are given a speaker-independent (SI)
acoustic model that performs reasonably well on the speech of almost
all speakers in general.  Our goal is to learn a personalized
speaker-dependent (SD) acoustic model for each target speaker that achieves
optimal ASR performance on his/her own speech. This is achieved by
adapting the SI model to the speech of each target speaker.

The speaker adaptation task is more challenging than the other types of domain
adaptation tasks in that it has only access to \emph{very limited} adaptation data
from the target speaker and has no access to the source domain data, e.g.
speech from other general speakers. Moreover, a deep neural network (DNN)
based SI model, usually with a large number of parameters, can easily get
overfitted to the limited adaptation data. To
address this issue, transformation-based approaches are introduced in
\cite{neto1995speaker, lhn} to reduce the number of
learnable parameters by inserting a linear network to the input, output or
hidden layers of the SI model. In \cite{svd_xue_1,svd_zhao}, the
trainable parameters are further reduced by singular value decomposition (SVD) of
weight matrices of a neural network and perform adaptation on an inserted
square matrix between the two low-rank matrices. Moreover, i-vector
\cite{ivector_saon} and speaker-code \cite{sc_abdel, sc_xue} are widely
used as auxiliary features to a neural network for speaker adaptation.
Further, regularization-based approaches are proposed in \cite{kld_yu, meng2019conditional, map_huang, huang2015rapid} to regularize the neuron output distributions or
the model parameters of the SD model such that it does not stray too far
away from the SI model.  

In this work, we propose a novel
regularization-based approach for speaker adaptation, in which we use adversarial multi-task learning (MTL) to regularize the
distribution of the deep features (i.e., hidden representations) in an SD
DNN acoustic model such that it does not deviate too much from the deep
feature distribution in the SI DNN acoustic model. We call this method \emph{adversarial speaker adaptation (ASA)}. Recently, adversarial training has achieved great success in learning
generative models \cite{gan}. In speech area, it has been applied to
acoustic model adaptation \cite{grl_sun, dsn_meng}, noise-robust
\cite{grl_shinohara, grl_serdyuk, meng2018adversarial}, speaker-invariant
\cite{meng2018speaker, saon2017english, meng2019aadit} ASR, speech enhancement \cite{pascual2017segan, meng2018afm, meng2018cycle} and speaker verification \cite{wang2018unsupervised, meng2019asv} using gradient reversal layer \cite{grl_ganin}
or domain separation network \cite{dsn}.  In these works, adversarial MTL assists in learning a deep intermediate feature that is both senone-discriminative and domain-invariant. 

In ASA, we
introduce an auxiliary discriminator network to classify whether an input
deep feature is generated by an SD or SI acoustic model. By using a fixed SI
acoustic model as the reference, the discriminator
network is jointly trained with the SD acoustic model to simultaneously
optimize the primary task of minimizing the senone classification loss
and the secondary task of mini-maximizing the SD/SI discrimination loss on
the adaptation data. Through this adversarial MTL, senone-discriminative deep features are
learned in the SD model with a distribution that is similar to that of the SI model. With such a regularized and adapted deep feature, the SD model is expected
to achieve improved ASR performance on the test speech from the target
speaker.
As an extension, ASA can also be performed on the senone posteriors (ASA-SP) to regularize the output distribution of the SD model.

We perform speaker adaptation experiments on Microsoft short message (SMD)
dictation dataset with 2600 hours of live US English data for training.
ASA achieves up to 14.4\% and 
7.9\% relative word error rate (WER) improvements for supervised and
unsupervised adaptation, respectively, over an SI model trained on 2600 hours of speech.

\section{Adversarial Speaker Adaptation}
\label{sec:asa}

In speaker adaptation task, for a target speaker, we are given a sequence of
adaptation speech frames $\mathbf{X}=\{\mathbf{x}_{1}, \ldots,
\mathbf{x}_{T}\}, \mathbf{x}_t \in \mathbbm{R}^{r_x}, t=1, \ldots, T$ from
the target speaker and a sequence of senone labels $\mathbf{Y}=\{y_{1},
\ldots, y_{T}\}, y_t \in \mathbbm{R}$ aligned with $\mathbf{X}$. For
supervised adaptation, $\mathbf{Y}$ is generated by aligning the adaptation
data against the transcription using SI acoustic model while for
unsupervised adaptation, the adaptation data is first decoded using the SI
acoustic model and the one-best path of the decoding lattice is used as
$\mathbf{Y}$.

\begin{figure}[htpb!]
	\centering
	\includegraphics[width=0.85\columnwidth]{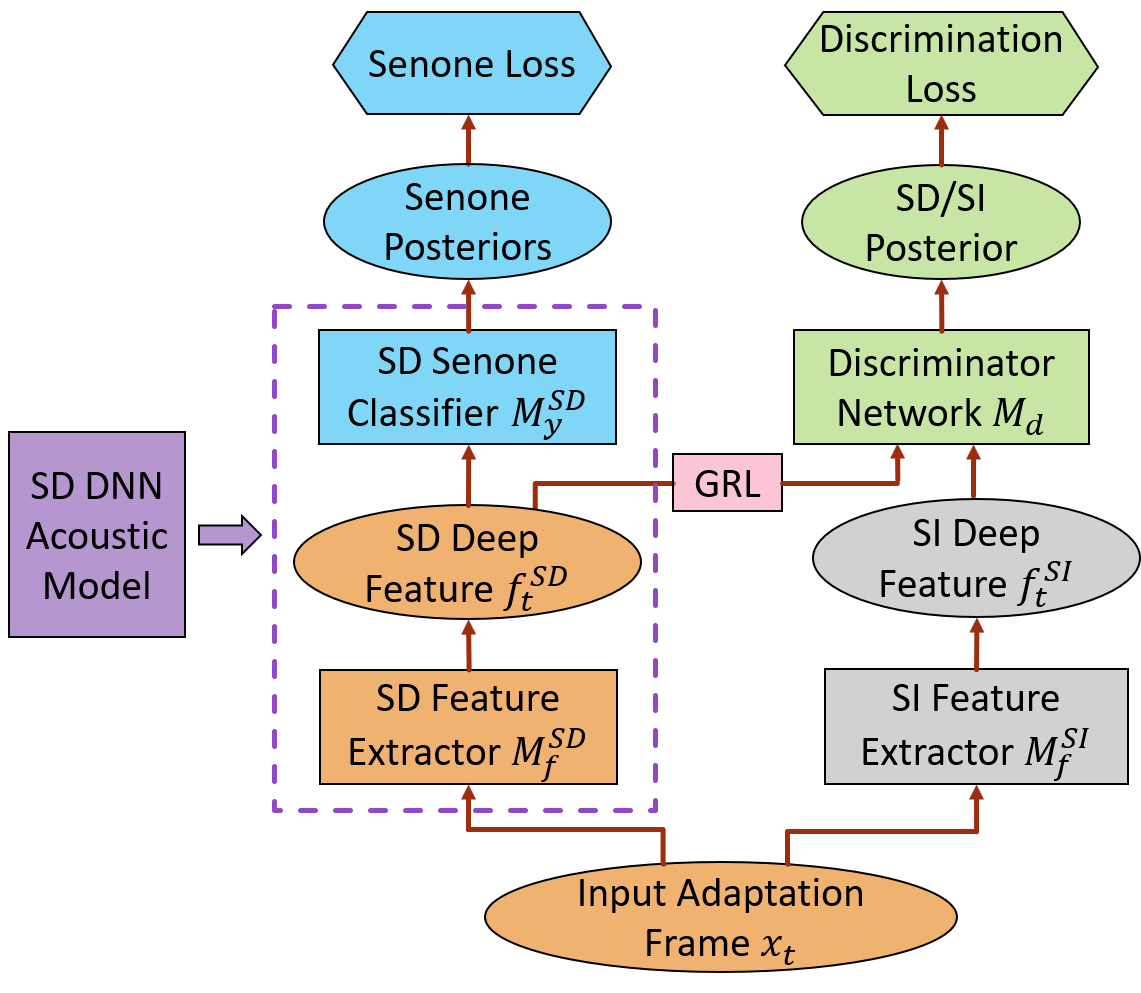}
	\caption{The framework of ASA. Only the
	optimized SD acoustic model consisting of $M_f^{\text{SD}}$ and
$M_y^{\text{SD}}$ are used for ASR on test data. $M_f^{\text{SI}}$ is fixed during ASA. $M_f^{\text{SI}}$ and $M_d$ are discarded after ASA.}
	\label{fig:asa}
\end{figure}

As shown in Fig. \ref{fig:asa}, we view the first few layers of a well-trained SI DNN acoustic model as an SI
feature extractor network $M_f^{\text{SI}}$ with parameters $\theta_f^{\text{SI}}$
and the the upper layers of the SI model as an SI senone classifier network $M_y^{\text{SI}}$ with parameters $\theta_y^{\text{SI}}$. $M_f^{\text{SI}}$
maps input adaptation speech frames $\mathbf{X}$ to
intermediate SI deep hidden features $\mathbf{F}^{\text{SI}}=\{\mathbf{f}_1^{\text{SI}}, \ldots,
\mathbf{f}_T^{\text{SI}}\}, \mathbf{f}_t^{\text{SI}}\in \mathbbm{R}^{r_f}$,
i.e.,
\begin{align}
	\mathbf{f}_t^{\text{SI}} = M_f^{\text{SI}}(\mathbf{x}_t), \footnotemark
	\label{eqn:fe_si}
\end{align}
\footnotetext{For recurrent DNN, $\mathbf{f}_t^{\text{SI}}$ also defends on $\{\mathbf{x}_1, \ldots, \mathbf{x}_{t-1}\}$ in addition to  $\mathbf{x}_t$ and we here simplify the notation to include only the current input. This abbreviation also applies to all the notations afterwards.}
and $M_y^{\text{SI}}$ with parameters
$\theta_y^{\text{SI}}$ maps $\mathbf{F}^{\text{SI}}$
to the posteriors $p(s|\mathbf{f}_t^{\text{SI}};
\theta_y^{\text{SI}})$ of a set of senones in $\mathbbm{S}$ as follows:
\begin{align}
	M_y^{\text{SI}}(\mathbf{f}_t^{\text{SI}}) = p(s | \mathbf{x}_t;
		\mathbf{\theta}^{\text{SI}}_f,
	\mathbf{\theta}^{\text{SI}}_y).
	\label{eqn:senone_classify}
\end{align}

An SD DNN acoustic model to be trained using speech from the target speaker is initialized from the
SI acoustic model. Specifically,  $M_f^{\text{SI}}$ is used to initialize SD feature extractor $M_f^{\text{SD}}$ with
parameters $\theta_f^{\text{SD}}$ and $M_y^{\text{SI}}$ is used to initialize SD senone classifier $M_y^{\text{SD}}$ with
parameters $\theta_y^{\text{SD}}$. Similarly, in an SD model,
$M_f^{\text{SD}}$ maps $\mathbf{x}_t$ to SD deep features
$\mathbf{f}_t^{\text{SD}}$ and $M_y^{\text{SD}}$ further transforms
$\mathbf{f}_t^{\text{SD}}$ to the same set of senone posteriors
$p(s|\mathbf{f}_t^{\text{SD}};
\theta_y^{\text{SD}}), s\in \mathbbm{S}$ as follows
\begin{align}
	M_y^{\text{SD}}(\mathbf{f}_t^{\text{SD}}) = M_y^{\text{SD}}(M_f^{\text{SD}}(\mathbf{x}_t)) = p(s | \mathbf{x}_t;
		\mathbf{\theta}^{\text{SD}}_f,
	\mathbf{\theta}^{\text{SD}}_y).
	\label{eqn:sd}
\end{align}
To adapt the SI model to target speech $\mathbf{X}$, we re-train the SD model 
by minimizing the cross-entropy senone classification loss between the predicted senone
posteriors and the senone labels $\mathbf{Y}$ below
\begin{align}
	&\mathcal{L}_{\text{senone}}(\theta_f^{\text{SD}}, \theta_y^{\text{SD}}) = -
	\frac{1}{T}\sum_{t = 1}^{T} \log p(y_t |
	\mathbf{x}_t;\theta_f^{\text{SD}}, \theta_y^{\text{SD}}) \nonumber \\[-5pt]
	&\qquad \quad =-\frac{1}{T}\sum_{t = 1}^T \sum_{s\in
		\mathbbm{S}} \mathbbm{1}[s =
	y_t] \log M_y(M_f^{\text{SD}}(\mathbf{x}_t)),
	\label{eqn:loss_m}
\end{align}
where $\mathbbm{1}[\cdot]$ is the indicator function which equals to 1 if the condition in the squared bracket is satisfied and 0 otherwise.

However, the adaptation data $\mathbf{X}$ is usually very limited for the
target speaker and the SI model with a large number of parameters can easily
get overfitted to the adaptation data. 
Therefore, we need to force the distribution of deep
hidden features $\mathbf{F}^{\text{SD}}$ in the SD model
 to be close to that of the deep features $\mathbf{F}^{\text{SI}}$ in SI model 
while minimizing $\mathcal{L}_{\text{senone}}$ as
follows
\begin{numcases}{}
	p({\mathbf{F}^{\text{SD}}}| \mathbf{X}, \theta_f^{\text{SD}}) \rightarrow
	p({\mathbf{F}^{\text{SI}}}| \mathbf{X}, \theta_f^{\text{SI}}) \label{eqn:asa_goal_1}, \\
	\min_{\theta_f^{\text{SD}}, \theta_y^{\text{SD}}} \mathcal{L}_{\text{senone}}(\theta_f^{\text{SD}}, \theta_y^{\text{SD}}).
	\label{eqn:asa_goal_2}
\end{numcases}
In KLD adaptation \cite{kld_yu},
the senone distribution estimated from the SD model is forced to be close to
that estimated from an SI model by adding  a KLD regularization to the
adaptation criterion. However, KLD is a distribution-wise \emph{asymmetric} measure which does not serve as a perfect distance metric between distributions \cite{kullback1951information}. For example, in \cite{kld_yu}, the minimization of $\mathcal{KL}(p^{\text{SI}} || p^{\text{SD}})$ does not guarantee $\mathcal{KL}(p^{\text{SD}} || p^{\text{SI}})$ is also minimized. In some cases, $\mathcal{KL}(p^{\text{SD}} || p^{\text{SI}})$ even increases as $\mathcal{KL}(p^{\text{SI}} || p^{\text{SD}})$ becomes smaller \cite{kurt2017kullback, moran2017kullback}. 
In ASA, we use adversarial MTL instead to push the distribution of $\mathbf{F}^{\text{SD}}$ towards 
that of $\mathbf{F}^{\text{SI}}$ while
being adapted to the target speech since the adversarial learning can guarantee that the global optimum is achieved if and only if $\mathbf{F}^{\text{SD}}$ and $\mathbf{F}^{\text{SI}}$ share exactly the same distribution \cite{gan}. 



To achieve Eq. \eqref{eqn:asa_goal_1}, we introduce an additional discriminator network $M_d$
with parameters $\theta_d$ which takes $\mathbf{F}^{\text{SD}}$ and
$\mathbf{F}^{\text{SI}}$ as the input and outputs the posterior probability
that an input deep feature is generated by the SD model, i.e.,
\begin{align}
	M_d(\mathbf{f}_t^{\text{SD}}) &= p(\mathbf{f}_t^{\text{SD}} \in
	\mathbbm{D}_{\text{SD}} | \mathbf{x}_t; \theta^{\text{SD}}_f, \theta_d ), \label{eqn:disc_sd} \\
	M_d(\mathbf{f}_t^{\text{SI}}) &= 1 - p(\mathbf{f}_t^{\text{SI}} \in
	\mathbbm{D}_{\text{SI}}| \mathbf{x}_t; \theta^{\text{SI}}_f, \theta_d ),
	\label{eqn:disc_si}
\end{align}
where $\mathbbm{D}_{\text{SD}}$ and $\mathbbm{D}_{\text{SI}}$ denote the
sets of SD and SI deep features, respectively.  The
discrimination loss $\mathcal{L}_{\text{disc}}(\theta_f, \theta_d)$ for
$M_d$ is formulated below using cross-entropy: 
\begin{align}
 \mathcal{L}_{\text{disc}}(\theta^{\text{SD}}_f,
	\theta^{\text{SI}}_f, \theta_d) = - \frac{1}{T}\sum_{t = 1}^{T} \left[ \log
		p(\mathbf{f}_t^{\text{SD}} \in \mathbbm{D}_{\text{SD}} |
	\mathbf{x}_t; \theta^{\text{SD}}_f, \theta_d) \right.\nonumber \\
	\qquad \qquad \qquad \qquad  \qquad \quad \left. + \log
	p(\mathbf{f}_t^{\text{SI}} \in \mathbbm{D}_{\text{SI}} |
\mathbf{x}_t; \theta^{\text{SI}}_f, \theta_d) \right] \nonumber \\
= - \frac{1}{T}\sum_{t = 1}^{T} \left\{\log M_d(M_f^{\text{SD}}(\mathbf{x}_t)) + \log
\left[1 - M_d(M_f^{\text{SI}}(\mathbf{x}_t))\right]\right\} \label{eqn:loss_d}
\end{align}
To make the distribution of $\mathbf{F}^{\text{SD}}$ similar to that of
$\mathbf{F}^{\text{SI}}$, we perform adversarial training of
$M_f^{\text{SD}}$ and $M_d$, i.e, we minimize $\mathcal{L}_{\text{disc}}$
with respect to $\theta_d$ and maximize $\mathcal{L}_{\text{disc}}$ with
respect to $\theta^{\text{SD}}_f$.  This minimax competition will first
increase the capability of $M_f^{\text{SD}}$ to generate
$\mathbf{F}^{\text{SD}}$ with a distribution similar to that of
$\mathbf{F}^{\text{SI}}$ and increase the discrimination capability of $M_d$. It
will eventually converge to the point where $M_f^{\text{SD}}$ generates
extremely confusing $\mathbf{F}^{\text{SD}}$ that $M_d$ is unable to
distinguish whether it is generated by $M_f^{\text{SD}}$ or
$M_f^{\text{SI}}$. At this point, we have successfully regularized the SD
model such that it does not deviate too much from the SI model and generalizes well to the test speech from target speaker.

With ASA, we want to learn a senone-discriminative SD deep feature with a similar
distribution to the SI deep features as in Eq. \eqref{eqn:asa_goal_1} and \eqref{eqn:asa_goal_2}. To achieve this, we perform
adversarial MTL, in which the SD model and $M_d$ 
are trained to jointly optimize the primary task of senone
classification and the secondary task of SD/SI discrimination with an
adversarial objective function as follows
\begin{align}
\hspace{-4pt} (\hat{\mathbf{\theta}}_f^{\text{SD}},
	\hat{\mathbf{\theta}}_y^{\text{SD}}) & =
	\argmin_{\mathbf{\theta}_f^{\text{SD}}, \mathbf{\theta}_y^{\text{SD}}}
	\mathcal{L}_{\text{senone}}(\mathbf{\theta}_f^{\text{SD}}, \mathbf{\theta}_y^{\text{SD}}) - 
	\lambda\mathcal{L}_{\text{disc}}(\mathbf{\theta}_f^{\text{SD}}, \mathbf{\theta}_f^{\text{SI}},
	 \hat{\mathbf{\theta}}_d),
     \label{eqn:minimax_1} \\
     (\hat{\mathbf{\theta}}_d) & =
     \argmin_{\mathbf{\theta}_d}
     \mathcal{L}_{\text{disc}}(\hat{\mathbf{\theta}}_f^{\text{SD}},\mathbf{\theta}_f^{\text{SI}},
 	\mathbf{\theta}_d), \label{eqn:minimax_2} 
\end{align}
where $\lambda$ controls the trade-off between
$\mathcal{L}_{\text{senone}}$ and $\mathcal{L}_{\text{disc}}$, and
$\hat{\mathbf{\theta}}_y^{\text{SD}}, \hat{\mathbf{\theta}}_f^{\text{SD}}$
and $\hat{\mathbf{\theta}}_d$ are the optimized parameters. Note that the
SI model serves only as a reference in ASA and its 
parameters $\mathbf{\theta}_y^{\text{SI}},
\mathbf{\theta}_f^{\text{SI}}$ are \emph{fixed} throughout the
optimization procedure.

The parameters are updated as follows via back propagation with stochastic gradient descent:
\vspace{-5pt}
\begin{align}
	& \mathbf{\theta}_f^{\text{SD}} \leftarrow \mathbf{\theta}_f^{\text{SD}} - \mu \left[ \frac{\partial
		\mathcal{L}_{\text{senone}}}{\partial \mathbf{\theta}_f^{\text{SD}}} - \lambda \frac{\partial
			\mathcal{L}_{\text{disc}}}{\partial
			\mathbf{\theta}_f^{\text{SD}}}
		\right],
		\label{eqn:grad_f} \\
	& \mathbf{\theta}_d \leftarrow \mathbf{\theta}_d - \mu \frac{\partial
		\mathcal{L}_{\text{disc}}}{\partial \mathbf{\theta}_d},
		\label{eqn:grad_s} \\
	& \mathbf{\theta}_y^{\text{SD}} \leftarrow \mathbf{\theta}_y^{\text{SD}} - \mu \frac{\partial
		\mathcal{L}_{\text{senone}}}{\partial \mathbf{\theta}_y^{\text{SD}}},
	\label{eqn:grad_y}
\end{align}
where $\mu$ is the learning rate.
For easy
implementation, gradient reversal layer is introduced in
\cite{grl_ganin}, which acts as
an identity transform in the forward propagation and multiplies the gradient
by $-\lambda$ during the backward propagation. Note that only the optimized SD DNN acoustic model consisting of $M_f^{\text{SD}}$ and
$M_y^{\text{SD}}$ is used for ASR on test data. $M_d$ and SI model are discarded after ASA. 

The procedure of ASA can be summarized in the steps below:
\begin{enumerate}
    \item Divide a well-trained and \emph{fixed} SI model into a feature extractor $M_f^{\text{SI}}$ followed by a senone classifier $M_y^{\text{SI}}$.
    \item Initialize the SD model with the SI model, i.e., clone $M_f^{\text{SD}}$ and $M_y^{\text{SD}}$ from $M_f^{\text{SI}}$ and $M_y^{\text{SI}}$, respectively.
    \item Add an auxiliary discriminator network $M_d$ taking SD and SI deep features, $\mathbf{F}^{\text{SD}}$ and $\mathbf{F}^{\text{SI}}$, as the input and predict the posterior that the input is generated by $M_f^{\text{SD}}$. 
    \item Jointly optimize $M_f^{\text{SD}}$, $M_y^{\text{SD}}$ and $M_d$ with adaptation data of a target speaker via adversarial MTL 
    as in Eq. \eqref{eqn:minimax_1} to \eqref{eqn:grad_y}.
    \item Use the optimized SD model consisting of $M_f^{\text{SD}}$, $M_y^{\text{SD}}$ for ASR decoding on test data of this target speaker.
\end{enumerate}

\section{Adversarial Speaker Adaptation on Senone Posteriors}
\label{sec:asa_senone}
\begin{figure}[htpb!]
	\centering
	\includegraphics[width=0.57\columnwidth]{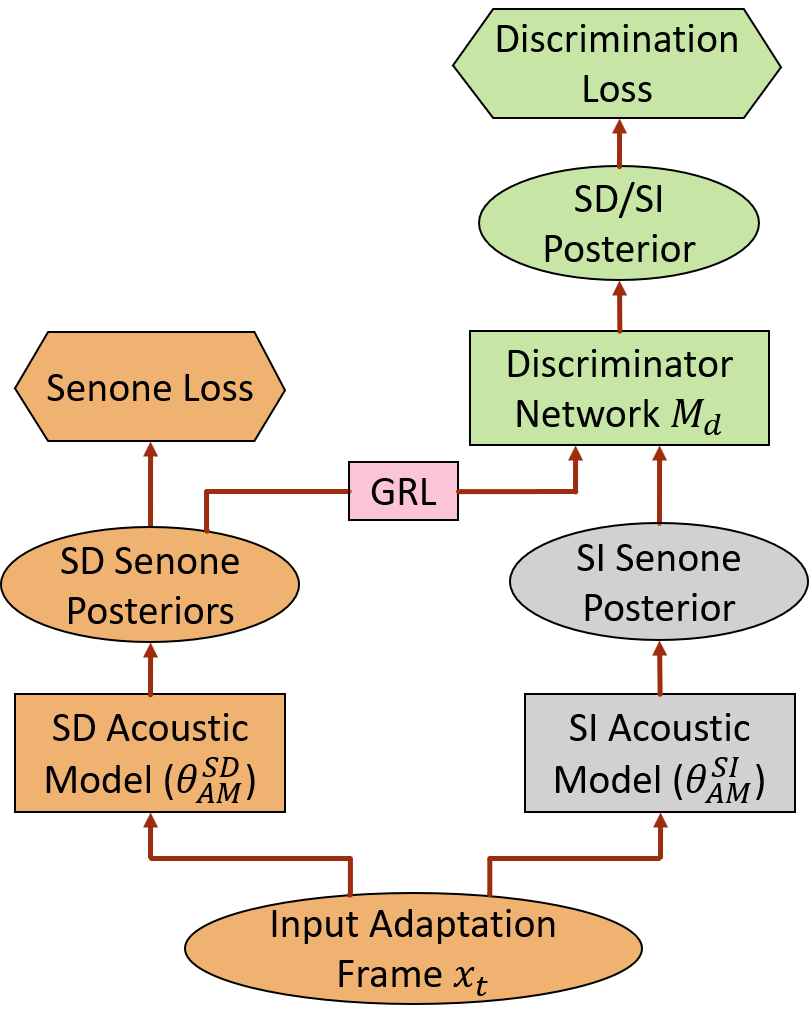}
	\vspace{-6pt}
	\caption{The framework of ASA-SP. The SI acoustic model is fixed during ASA-SP. Only the SD acoustic model is used in ASR.}
	\label{fig:asa_senone}
\vspace{-10pt}
\end{figure}

As shown in Fig. \ref{fig:asa_senone}, in ASA-SP, adversarial learning is applied to regularize the vectors of senone posteriors $\mathbf{y}^{\text{SD}}_t = \left[p(s|\mathbf{x}_t; \mathbf{\theta}^{\text{SD}}_{\text{AM}})\right]_{s\in\mathbbm{S}}$ predicted by the SD model
to be close to that of a well-trained and \emph{fixed} SI model, i.e., $\mathbf{y}^{\text{SI}}_t = \left[p(s|\mathbf{x}_t; \mathbf{\theta}^{\text{SI}}_{\text{AM}})\right]_{s\in\mathbbm{S}}$ while simultaneously minimizing the senone loss $\mathcal{L}_{\text{senone}}(\mathbf{\theta}^{\text{SD}}_{\text{AM}})$, where  $\mathbf{\theta}^{\text{SD}}_{\text{AM}} = \{\mathbf{\theta}^{\text{SD}}_f, \mathbf{\theta}^{\text{SD}}_y\}$ and $\mathbf{\theta}^{\text{SI}}_{\text{AM}} = \{\mathbf{\theta}^{\text{SI}}_f, \mathbf{\theta}^{\text{SI}}_y\}$ are SI and SD model parameters, respectively.

In this case, the discriminator $M_d$ takes $p(s|\mathbf{x}_t; \mathbf{\theta}^{\text{SD}}_{\text{AM}})$ and $p(s|\mathbf{x}_t; \mathbf{\theta}^{\text{SI}}_{\text{AM}})$ as the input and predicts the posterior that the input is generated by the SD model. The
discrimination loss $\mathcal{L}_{\text{disc}}(\theta_f, \theta_d)$ for
$M_d$ is formulated below using cross-entropy:
\begin{align}
 & \mathcal{L}_{\text{disc}}(\theta^{\text{SD}}_{\text{AM}},
	\theta_d) = - \frac{1}{T}\sum_{t = 1}^{T} \left[ \log
		p(\mathbf{y}_t^{\text{SD}} \in \mathbbm{E}_{\text{SD}} |
	\mathbf{x}_t; \theta^{\text{SD}}_{\text{AM}}, \theta_d) \right.\nonumber \\
	& \qquad \qquad \qquad \qquad  \qquad \left. + \log
	p(\mathbf{y}_t^{\text{SI}} \in \mathbbm{E}_{\text{SI}} |
\mathbf{x}_t; \theta^{\text{SI}}_{\text{AM}}, \theta_d) \right],
\label{eqn:loss_d_2}
\end{align}
where $\mathbbm{E}_{\text{SD}}$ and $\mathbbm{E}_{\text{SI}}$ denote the
sets of senone posterior vectors generated by SD and SI models, respectively. Similar adversarial MTL is performed to make the distributions of $\mathbf{Y}^{\text{SD}} = \{\mathbf{y}^{\text{SD}}_1, \ldots, \mathbf{y}^{\text{SD}}_T\}$ similar to that of $\mathbf{Y}^{\text{SI}} = \{\mathbf{y}^{\text{SI}}_1, \ldots, \mathbf{y}^{\text{SI}}_T\}$ below:
\vspace{-5pt}
\begin{align}
	(\hat{\mathbf{\theta}}_{\text{AM}}^{\text{SD}}) & =
	\argmin_{\mathbf{\theta}_{\text{AM}}^{\text{SD}}}
	\mathcal{L}_{\text{senone}}(\mathbf{\theta}_{\text{AM}}^{\text{SD}}) - 
	\lambda\mathcal{L}_{\text{disc}}(\mathbf{\theta}_{\text{AM}}^{\text{SD}}, \mathbf{\theta}_{\text{AM}}^{\text{SI}},
	 \hat{\mathbf{\theta}}_d),
     \label{eqn:minimax_3} \\
     (\hat{\mathbf{\theta}}_d) & =
     \argmin_{\mathbf{\theta}_d}
     \mathcal{L}_{\text{disc}}(\hat{\mathbf{\theta}}_{\text{AM}}^{\text{SD}},\mathbf{\theta}_{\text{AM}}^{\text{SI}},
 	\mathbf{\theta}_d). \label{eqn:minimax_4}
\end{align}
$\mathbf{\theta}_{\text{AM}}^{\text{SI}}$ is optimized by back propagation below, $\mathbf{\theta}_d$ is optimized via Eq. \eqref{eqn:grad_s} and $\mathbf{\theta}_{\text{AM}}^{\text{SI}}$ remain unchanged during the optimization.
\begin{align}
	& \mathbf{\theta}_{\text{AM}}^{\text{SD}} \leftarrow \mathbf{\theta}_{\text{AM}}^{\text{SD}} - \mu \left[ \frac{\partial
		\mathcal{L}_{\text{senone}}}{\partial \mathbf{\theta}_f^{\text{SD}}} - \lambda \frac{\partial
			\mathcal{L}_{\text{disc}}}{\partial
			\mathbf{\theta}_{\text{AM}}^{\text{SD}}}
		\right]. \label{eqn:grad_f_2}
\end{align}

ASA-SP is an extension of ASA where the deep hidden feature moves up to the output layer and becomes the senone posteriors vector. In this case, the senone classifier disappears and the feature extractor becomes the entire acoustic model.


\section{Experiments}
We perform speaker adaptation on a Microsoft Windows Phone
SMD task. The training data consists of 2600
hours of Microsoft internal live US English data collected through a number
of deployed speech services including voice search and SMD. The test set
consists of 7 speakers with a total number of 20,203 words.  Four
adaptation sets of 20, 50, 100 and 200 utterances per speaker are used for acoustic model adaptation, respectively, to explore the impact of adaptation data duration. Any adaptation set with smaller number of
utterances is a subset of a larger one.  

\vspace{-4pt}
\subsection{Baseline System}
We train an SI long short-term memory (LSTM)-hidden Markov model (HMM)
acoustic model \cite{sak2014long, meng2017deep, erdogan2016multi} with 2600 hours of training data.  This SI model has 4
hidden layers with 1024 units in each layer and the output size of
each hidden layer is reduced to 512 by a linear projection. 80-dimensional
log Mel filterbank features are extracted from training, adaptation and
test data. The output layer has a dimension of 5980. The LSTM is trained to
minimize the frame-level cross-entropy criterion. There is no frame
stacking, and the output HMM state label is delayed by 5 frames. A trigram
LM is used for decoding with around 8M n-grams. This SI LSTM acoustic model
achieves 13.95\% WER on the SMD test set.


We further perform KLD speaker adaptation \cite{kld_yu} with different regularization weights $\rho$. In Tables \ref{table:asa_sup} and
\ref{table:asa_uns}, KLD with $\rho =
0.5$ achieves 12.54\% - 13.24\% and 13.55\% - 13.85\% WERs 
for supervised and unsupervised adaption,
respectively with 20 - 200 adaptation utterances.

\begin{table}[h]
\setlength{\tabcolsep}{5.5 pt}
\centering
\begin{tabular}[c]{c|c|c|c|c|c}
	\hline
	\hline
	\multirow{2}{*}{\begin{tabular}{@{}c@{}} System 
		\end{tabular}} & \multicolumn{5}{c}{Number of Adaptation Utterances} \\
	\hhline{~-----}	
	& 20 & 50 & 100 & 200 & Avg. \\
	\hline
	SI & \multicolumn{5}{c}{13.95} \\
	\hline
	KLD ($\rho=0.0$) & 13.68 & 13.39 & 13.31 & 13.21 & 13.40 \\ 
	\hline
	KLD ($\rho=0.2$) & 13.20 & 13.00 & 12.71 & 12.61 & 12.88  \\
	\hline
	KLD ($\rho=0.5$) & 13.24 & 13.08 & 12.85 & 12.54 & 12.93 \\
	\hline
	KLD ($\rho=0.8$) & 13.55 & 13.50  & 13.46 & 13.17 & 13.42 \\
	\hline
	\hline
	ASA ($\lambda=1.0$) & \textbf{12.99} & 12.86 & 12.56 & 12.05 & 12.62 \\
	\hline
	ASA ($\lambda=3.0$) & 13.03 & 12.72 & \textbf{12.35} & \textbf{11.94} &
	\textbf{12.56} \\
	\hline
	ASA ($\lambda=5.0$) & 13.14 & \textbf{12.71} & 12.50 & 12.06 & 12.60 \\
	\hline
	\hline
	ASA-SP ($\lambda=1.0$) & 13.04 & 12.88 & 12.66 & 12.17 & 12.69 \\
	\hline
	ASA-SP ($\lambda=3.0$) & 13.05 & 12.89 & 12.67 & 12.18 & 12.70 \\
	\hline
	ASA-SP ($\lambda=5.0$) & 13.05 & 12.89 & 12.69 & 12.18 & 12.70 \\
	\hline
	\hline
\end{tabular}
\caption{The WER (\%) performance of \textbf{supervised} speaker adaptation
		using KLD 
	and ASA with different $\lambda$ on Microsoft SMD task. }
\label{table:asa_sup}
\end{table}
\vspace{-10pt}

\begin{table}[h]
\setlength{\tabcolsep}{5.5pt}
\centering
\begin{tabular}[c]{c|c|c|c|c|c}
	\hline
	\hline
	\multirow{2}{*}{\begin{tabular}{@{}c@{}} System 
		\end{tabular}} & \multicolumn{5}{c}{Number of Adaptation Utterances} \\
	\hhline{~-----}	
	& 20 & 50 & 100 & 200 & Avg. \\
	\hline
	SI  & \multicolumn{5}{c}{13.95} \\
	\hline
	KLD ($\rho=0.0$) & 14.18 & 14.01 & 13.81 & 13.73 & 13.93 \\
	\hline
	KLD ($\rho=0.2$) & 13.86 & 13.83 & 13.75 & 13.65 & 13.77 \\
	\hline
	KLD ($\rho=0.5$) & 13.85 & 13.80 & 13.73 & 13.55 & 13.73 \\
	\hline
	KLD ($\rho=0.8$) & 13.89 & 13.86 & 13.80 & 13.72 & 13.82 \\
	\hline
	\hline
	ASA ($\lambda=1.0$) & 13.74 & 13.70 & 13.38 & 12.99 & 13.45 \\
	\hline
	ASA ($\lambda=3.0$) & \textbf{13.66} & \textbf{13.61} & \textbf{13.09} & \textbf{12.85} & \textbf{13.30} \\
	\hline
	ASA ($\lambda=5.0$) & 13.85 & 13.69 & 13.27 & 13.03 & 13.46 \\
	\hline
	\hline
	ASA-SP ($\lambda=1.0$) & 13.83	& 13.72 & 13.48 & 13.16 & 13.55 \\
	\hline
	ASA-SP ($\lambda=3.0$) & 13.84	& 13.74 & 13.53 & 13.17 & 13.57 \\
	\hline
	ASA-SP ($\lambda=5.0$) & 13.85 & 13.74 & 13.52 & 13.16 & 13.57 \\
	\hline
	\hline
\end{tabular}
\caption{The WER (\%) performance of \textbf{unsupervised} speaker adaptation
		using KLD 
	and ASA with different $\lambda$ on Microsoft SMD task. }
\label{table:asa_uns}
\end{table}
\vspace{-15pt}

\subsection{Adversarial Speaker Adaptation}
\label{sec:asa}
We perform standard ASA as described in Section \ref{sec:asa}.  The
SI feature extractor $M_f^{\text{SI}}$ is formed as the first $N_h$ layers
of the SI LSTM and the SI senone classifier $M_y^{\text{SI}}$ is the rest
$(4-N_h)$ hidden layers plus the output layer. The SD feature extractor
$M_f^{\text{SD}}$ and SD senone classifier $M_y^{\text{SD}}$ are cloned
from $M_f^{\text{SI}}$ and $M_y^{\text{SI}}$, respectively as an
initialization. $N_h$ indicates the position of the deep hidden feature in
the SD and SI LSTMs.  $M_d$ is a
feedforward DNN with 2 hidden layers and 512 hidden units for each layer.
The output layer of $M_d$ has 1 unit predicting the posteriors of input
deep feature generated by the $M_f^{\text{SD}}$.  $M_d$ has 512-dimensional
input layer.  $M_f^{\text{SD}}$, $M_y^{\text{SD}}$ and $M_d$ are jointly
trained with an adversarial MTL objective. Due to space limitation, we only show the results when $N_h=4$. \footnote{It has been shown in \cite{dsn_meng, grl_sun} that the ASR performance increases with the growth of $N_h$ for adversarial domain adaptation. The same trend is observed for ASA experiments.}


For supervised ASA, the same alignment is used as in KLD. 
In Table \ref{table:asa_sup}, the best ASA setups achieve 
12.99\%, 12.71\%, 12.35\% and 11.94\% WERs for 20, 50, 100, 200 adaptation
utterances which improve the WERs by 6.9\%, 8.9\%, 11.5\%, 14.4\%
relatively over the SI LSTM, respectively. 
Supervised ASA ($\lambda=3.0$) also achieves up to 5.3\% relative WER reduction over the best KLD setup ($\rho=0.2$).  

For unsupervised ASA, the same decoded senone labels are used as in KLD.
In Table \ref{table:asa_uns}, the best ASA setups
achieve 13.66\%, 13.61\%, 13.09\% and 12.85\% WERs for 20, 50, 100, 200
adaptation utterances which improves the WERs by 2.1\%, 2.4\%, 6.2\%, 7.9\%
relatively over the SI LSTM, respectively. 
Unsupervised ASA ($\lambda=3.0$) also achieves up to and 5.2\%
relatively WER gains over the best KLD setup ($\rho=0.5$).
Compared with
supervised ASA, the unsupervised one decreases the relative WER gain
over the SI LSTM by about half on the same number of adaptation utterances.

For both supervised and unsupervised ASA, the WER first decreases as $\lambda$ grows larger and then increases when $\lambda$ becomes too large. 
ASA
performs consistently better than SI LSTM and KLD with different number of
adaptation utterances for both supervised and unsupervised adaptation.\footnote{In this work, we only compare ASA with the most popular regularization-based approach, i.e., KLD, because the other approaches such as transformation, SVD, auxiliary feature, etc.
are orthogonal to ASA and can be used together with ASA to get additional WER improvement.} The relative
gain increases as the number of adaptation utterance grows.

\vspace{-5pt} 
\subsection{Adversarial Speaker Adaptation on Senone Posteriors}
We then perform ASA-SP as described in Section \ref{sec:asa}. The SD acoustic model is cloned
from the SI LSTM as the initialization. $M_d$ shares the same architecture as the one in Section \ref{sec:asa}.
In Table \ref{table:asa_sup}, for supervised adaptation ASA-SP $(\lambda = 1.0)$ achieves 6.5\%, 7.7\%, 9.2\%, 12.8\% relative WER gain
over the SI LSTM, respectively and up to 3.5\% relative WER reduction over the best KLD setup ($\rho=0.2$). 
In Table \ref{table:asa_uns}, for unsupervised adaptation ASA-SP $(\lambda = 1.0)$ achieves 0.9\%, 1.6\%, 3.4\%, 5.7\%, 2.9\% relative WER gain
over the SI LSTM, respectively and up to 4.8\% relative WER reduction over the best KLD setup ($\rho=0.5$). 

Although ASA-SP consistently improves over KLD on different number of
adaptation utterances for both supervised and unsupervised adaptation, it performs worse than standard ASA where the regularization from SI model is performed at the hidden layers. The reason is that the senone posteriors vectors $\mathbf{y}^{\text{SI}}_t$, $\mathbf{y}^{\text{SD}}_t$ lie in a much higher-dimensional space than the deep features $\mathbf{f}^{\text{SI}}_t$, $\mathbf{f}^{\text{SD}}_t$ so that the discriminator is much harder to learn given much sparser-distributed samples. We also notice that ASA-SP performance is much less sensitive to the variation of $\lambda$ compared with standard ASA.

\vspace{-3pt} 
\section{Conclusion}
In this work, a novel adversarial speaker adaptation method is proposed, in
which the deep hidden features (ASA) or the output senone posteriors (ASA-SP) of an SD DNN acoustic model are forced by
the adversarial MTL to conform to a similar distribution as those
of a fixed reference SI DNN acoustic model while being trained to be
senone-discriminative with the limited adaptation data.

We evaluate ASA on Microsoft SMD task with 2600 hours of training data. 
ASA achieves up to 14.4\% and 7.9\% relative WER gain for supervised and unsupervised adaptation, respectively, over the SI LSTM acoustic model.
ASA also improves consistently over the KLD
regularization method. The relative gain grows as the number of adaption utterances increases. ASA-SP performs consistently better than KLD but worse than the standard ASA.

\vfill\pagebreak

\bibliographystyle{IEEEbib}
\bibliography{refs}

\begin{thebibliography}{10}

\bibitem{DNN4ASR-hinton2012}
G.~Hinton, L.~Deng, D.~Yu, et~al.,
\newblock ``Deep neural networks for acoustic modeling in speech recognition:
  The shared views of four research groups,''
\newblock {\em IEEE Signal Processing Magazine}, vol. 29, no. 6, pp. 82--97,
  2012.

\bibitem{yu2017recent}
Dong Yu and Jinyu Li,
\newblock ``Recent progresses in deep learning based acoustic models,''
\newblock {\em IEEE/CAA Journal of Automatica Sinica}, vol. 4, no. 3, pp.
  396--409, 2017.

\bibitem{Li14overview}
J.~Li, L.~Deng, Y.~Gong, and R.~Haeb-Umbach,
\newblock ``An overview of noise-robust automatic speech recognition,''
\newblock {\em IEEE/ACM Transactions on Audio, Speech and Language Processing},
  vol. 22, no. 4, pp. 745--777, April 2014.

\bibitem{neto1995speaker}
J.~Neto, L.~Almeida, M.~Hochberg, et~al.,
\newblock ``Speaker-adaptation for hybrid hmm-ann continuous speech recognition
  system,''
\newblock in {\em Proc. Eurospeech}, 1995.

\bibitem{lhn}
R.~Gemello, F.~Mana, S.~Scanzio, et~al.,
\newblock ``Linear hidden transformations for adaptation of hybrid ann/hmm
  models,''
\newblock {\em Speech Communication}, vol. 49, no. 10, pp. 827 -- 835, 2007.

\bibitem{svd_xue_1}
Jian Xue, Jinyu Li, and Yifan Gong,
\newblock ``Restructuring of deep neural network acoustic models with singular
  value decomposition.,''
\newblock in {\em Interspeech}, 2013, pp. 2365--2369.

\bibitem{svd_zhao}
Y.~Zhao, J.~Li, and Y.~Gong,
\newblock ``Low-rank plus diagonal adaptation for deep neural networks,''
\newblock in {\em Proc. ICASSP}, March 2016, pp. 5005--5009.

\bibitem{ivector_saon}
G.~Saon, H.~Soltau, D.~Nahamoo, and M.~Picheny,
\newblock ``Speaker adaptation of neural network acoustic models using
  i-vectors,''
\newblock in {\em Proc. ASRU}, Dec 2013, pp. 55--59.

\bibitem{sc_abdel}
O.~Abdel-Hamid and H.~Jiang,
\newblock ``Fast speaker adaptation of hybrid nn/hmm model for speech
  recognition based on discriminative learning of speaker code,''
\newblock in {\em Proc. ICASSP}, May 2013, pp. 7942--7946.

\bibitem{sc_xue}
S.~Xue, O.~Abdel-Hamid, H.~Jiang, L.~Dai, and Q.~Liu,
\newblock ``Fast adaptation of deep neural network based on discriminant codes
  for speech recognition,''
\newblock {\em in TASLP}, vol. 22, no. 12, pp. 1713--1725, Dec 2014.

\bibitem{kld_yu}
D.~Yu, K.~Yao, H.~Su, G.~Li, and F.~Seide,
\newblock ``Kl-divergence regularized deep neural network adaptation for
  improved large vocabulary speech recognition,''
\newblock in {\em Proc. ICASSP}, May 2013, pp. 7893--7897.

\bibitem{meng2019conditional}
Zhong Meng, Jinyu Li, Yong Zhao, and Yifan Gong,
\newblock ``Conditional teacher-student learning,''
\newblock in {\em Proc. ICASSP}, 2019.

\bibitem{map_huang}
Z.~Huang, S.~Siniscalchi, I.~Chen, et~al.,
\newblock ``Maximum a posteriori adaptation of network parameters in deep
  models,''
\newblock in {\em Proc. Interspeech}, 2015.

\bibitem{huang2015rapid}
Z.~Huang, J.~L, S.~Siniscalchi., et~al.,
\newblock ``Rapid adaptation for deep neural networks through multi-task
  learning.,''
\newblock in {\em Interspeech}, 2015, pp. 3625--3629.

\bibitem{gan}
I.~Goodfellow, J.~Pouget-Adadie, et~al.,
\newblock ``Generative adversarial nets,''
\newblock in {\em Proc. NIPS}, pp. 2672--2680. 2014.

\bibitem{grl_sun}
S.~Sun, B.~Zhang, L.~Xie, et~al.,
\newblock ``An unsupervised deep domain adaptation approach for robust speech
  recognition,''
\newblock {\em Neurocomputing}, vol. 257, pp. 79 -- 87, 2017.

\bibitem{dsn_meng}
Z.~Meng, Z.~Chen, V.~Mazalov, J.~Li, and Y.~Gong,
\newblock ``Unsupervised adaptation with domain separation networks for robust
  speech recognition,''
\newblock in {\em Proc. ASRU}, 2017.

\bibitem{grl_shinohara}
Yusuke Shinohara,
\newblock ``Adversarial multi-task learning of deep neural networks for robust
  speech recognition.,''
\newblock in {\em INTERSPEECH}, 2016, pp. 2369--2372.

\bibitem{grl_serdyuk}
D.~Serdyuk, K.~Audhkhasi, P.~Brakel, B.~Ramabhadran, et~al.,
\newblock ``Invariant representations for noisy speech recognition,''
\newblock in {\em NIPS Workshop}, 2016.

\bibitem{meng2018adversarial}
Zhong Meng, Jinyu Li, Yifan Gong, and Biing-Hwang Juang,
\newblock ``Adversarial teacher-student learning for unsupervised domain
  adaptation,''
\newblock in {\em Proc. ICASSP}. IEEE, 2018, pp. 5949--5953.

\bibitem{meng2018speaker}
Z.~Meng, J.~Li, Z.~Chen, et~al.,
\newblock ``Speaker-invariant training via adversarial learning,''
\newblock in {\em Proc. ICASSP}, 2018.

\bibitem{saon2017english}
G.~Saon, G.~Kurata, T.~Sercu, et~al.,
\newblock ``English conversational telephone speech recognition by humans and
  machines,''
\newblock {\em arXiv preprint arXiv:1703.02136}, 2017.

\bibitem{meng2019aadit}
Zhong Meng, Jinyu Li, and Yifan Gong,
\newblock ``Attentive adversarial learning for domain-invariant training,''
\newblock in {\em Proc. ICASSP}, 2019.

\bibitem{pascual2017segan}
Santiago Pascual, Antonio Bonafonte, and Joan Serr{\`a},
\newblock ``Segan: Speech enhancement generative adversarial network,''
\newblock in {\em Interspeech}, 2017.

\bibitem{meng2018afm}
Zhong Meng, Jinyu Li, and Yifan Gong,
\newblock ``Adversarial feature-mapping for speech enhancement,''
\newblock {\em Interspeech}, 2018.

\bibitem{meng2018cycle}
Zhong Meng, Jinyu Li, and Yifan Gong,
\newblock ``Cycle-consistent speech enhancement,''
\newblock {\em Interspeech}, 2018.

\bibitem{wang2018unsupervised}
Q.~Wang, W.~Rao, S.~Sun, L.~Xie, E.~S. Chng, and H.~Li,
\newblock ``Unsupervised domain adaptation via domain adversarial training for
  speaker recognition,''
\newblock {\em ICASSP 2018}, 2018.

\bibitem{meng2019asv}
Zhong Meng, Yong Zhao, Jinyu Li, and Yifan Gong,
\newblock ``Adversarial speaker verification,''
\newblock in {\em Proc. ICASSP}, 2019.

\bibitem{grl_ganin}
Yaroslav Ganin and Victor Lempitsky,
\newblock ``Unsupervised domain adaptation by backpropagation,''
\newblock in {\em Proc. ICML}, Lille, France, 2015, vol.~37, pp. 1180--1189,
  PMLR.

\bibitem{dsn}
K.~Bousmalis, G.~Trigeorgis, N.~Silberman, et~al.,
\newblock ``Domain separation networks,''
\newblock in {\em Proc. NIPS}, D.~D. Lee, M.~Sugiyama, U.~V. Luxburg, I.~Guyon,
  and R.~Garnett, Eds., pp. 343--351. Curran Associates, Inc., 2016.

\bibitem{kullback1951information}
Solomon Kullback and Richard~A Leibler,
\newblock ``On information and sufficiency,''
\newblock {\em The annals of mathematical statistics}, vol. 22, no. 1, pp.
  79--86, 1951.

\bibitem{kurt2017kullback}
Will Kurt,
\newblock ``Kullback-leibler divergence explained,''
  \url{https://www.countbayesie.com/blog/2017/5/9/kullback-leibler-divergence-explained},
  2017.

\bibitem{moran2017kullback}
Ben Moran,
\newblock ``Kullback-leibler divergence asymmetry,''
  \url{https://benmoran.wordpress.com/2012/07/14/kullback-leibler-divergence-asymmetry/},
  2012.

\bibitem{sak2014long}
F.~Beaufays H.~Sak, A.~Senior,
\newblock ``Long short-term memory recurrent neural network architectures for
  large scale acoustic modeling,''
\newblock in {\em Interspeech}, 2014.

\bibitem{meng2017deep}
Z.~Meng, S.~Watanabe, J.~R. Hershey, et~al.,
\newblock ``Deep long short-term memory adaptive beamforming networks for
  multichannel robust speech recognition,''
\newblock in {\em ICASSP}. IEEE, 2017, pp. 271--275.

\bibitem{erdogan2016multi}
H.~Erdogan, T.~Hayashi, J.~R. Hershey, et~al.,
\newblock ``Multi-channel speech recognition: Lstms all the way through,''
\newblock in {\em CHiME-4 workshop}, 2016, pp. 1--4.

\end{thebibliography}

\end{document}